%% file: root.tex
\documentclass[letterpaper, 10 pt, conference]{ieeeconf}  

\IEEEoverridecommandlockouts                              

\usepackage{graphics} 
\usepackage{epsfig} 
\usepackage{mathptmx} 
\usepackage{times} 
\usepackage{amsmath} 
\usepackage{amssymb}  

\usepackage{xcolor}

\usepackage{booktabs} 
\usepackage{multirow}
\usepackage{siunitx}

\usepackage[abbreviations]{glossaries-extra}
\glssetcategoryattribute{acronym}{indexonlyfirst}{true}
\newabbreviation{spf}{SPF}{Semantic Pointcloud Filter}
\newabbreviation{ss}{SS}{support surface}
\newabbreviation{pc}{PC}{pointcloud}
\newabbreviation{rmse}{RMSE}{root-mean-square error}
\newabbreviation{GP}{GP}{Gaussian process}
\newabbreviation{RBF}{RBF}{Radial Basis Function}
\newabbreviation{ssde}{SSDE}{support surface depth estimation}
\newabbreviation{ssseg}{SSSeg}{support surface segmentation}
\newabbreviation{sgss}{SGSS}{self-generated semantic segmentation}
\newabbreviation{miou}{mIoU}{mean intersection of union}

\newabbreviation{fov}{FoV}{field of view}

\usepackage{graphicx}
\usepackage{todonotes}

\newcommand{\ra}[1]{\renewcommand{\arraystretch}{#1}}

\newcommand{\testvspace}{\vspace{-0.5cm}}

\title{\LARGE \bf
Seeing Through the Grass: \\Semantic Pointcloud Filter for Support Surface Learning

}
\author{Anqiao Li, Chenyu Yang, Jonas Frey$^{\dagger}$, Joonho Lee, Cesar Cadena, and Marco Hutter %
\thanks{All authors are with the Robotic Systems Lab, ETH Zurich, 8092 Zurich, Switzerland. $\texttt{\{anqiali}$, $\texttt{chenyang}$, $\texttt{jonfrey}$, $\texttt{jolee}$, $\texttt{cesarc}$, $\texttt{mhutter}$$\}$ $\texttt{@ethz.ch}$}%
\thanks{$\dagger$ The author is with the Max Planck ETH CLS.}%
\thanks{
This work was supported by the Swiss National Science Foundation (SNSF) through project 188596, the National Centre of Competence in Research Robotics (NCCR Robotics), the European Union's Horizon 2020 research and innovation program under grant agreement No 101016970, No 101070405, and No 852044, and an ETH Zurich Research Grant. Jonas Frey is supported by the Max Planck ETH Center for Learning Systems.
}
}

\begin{document}

\maketitle
\thispagestyle{empty}
\pagestyle{empty}

\input{chapters/0_abstract}

\section{INTRODUCTION}
\input{chapters/1_introduction}

\begin{figure*}[!t]
    \centering
    \includegraphics[width=\linewidth,trim=0cm 0cm 6cm 0cm,clip]{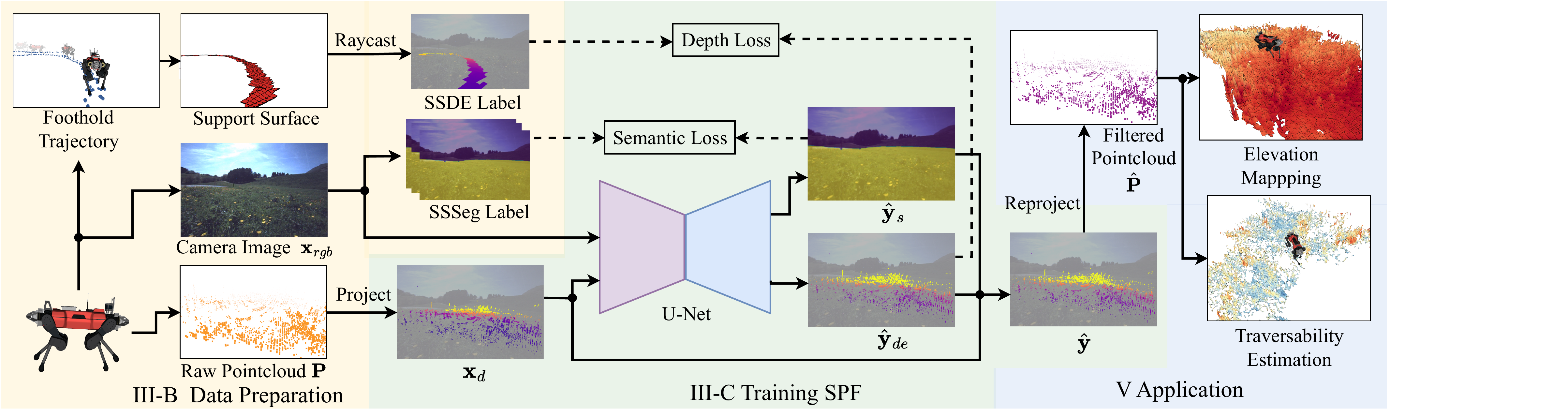}
    \caption{Overview of the system and corresponding sections. The training data is generated from pre-recorded data, consisting of raw pointcloud and camera images acquired from onboard sensors of the robot and robot trajectories obtained through exteroceptive observations. The support surface can be reconstructed from footholds extracted from the trajectories. 
    When training the SPF, the raw pointcloud data $P$ is first projected onto image space as $\mathbf{x}_d$ and concatenated with the camera image. 
    The U-net-based neural network predicts jointly the depth ( using supervision of support surface depth estimation (SSDE) labels) and semantic segmentation (using supervision of hand-labeled semantic segmentation (SSSeg) labels). The final output $\hat{\mathbf{y}}$ is given by combining $\mathbf{x}_d$ and the predicted depth estimation $\hat{\mathbf{y}}_{de}$ based on the predicted semantic segmentation mask $\hat{\mathbf{y}}_{s}$. The filtered pointcloud $\hat{P}$ is reprojected from $\hat{\mathbf{y}}$ into 3D and used for two downstream tasks: elevation mapping and traversability estimation. }
    \label{fig:system}
    \testvspace
\end{figure*}
\section{RELATED WORK}
\input{chapters/2_related_work}

\section{METHOD}
\input{chapters/3_method}

\section{IMPLEMENTATION DETAILS}
\input{chapters/4__Implementation_Details}

\section{EXPERIMENTS}
\input{chapters/4_experiments}

\section{CONCLUSION}

\input{chapters/5_conclusion}

\bibliographystyle{IEEEtran}
\bibliography{bibliography}{}

\end{document}

%% file: chapters/0_abstract.tex
\begin{abstract}
Mobile ground robots require perceiving and understanding their surrounding support surface to move around autonomously and safely. The support surface is commonly estimated based on exteroceptive depth measurements, e.g., from LiDARs. However, the measured depth fails to align with the true support surface in the presence of high grass or other penetrable vegetation. In this work, we present the \gls{spf}, a Convolutional Neural Network (CNN) that learns to adjust LiDAR measurements to align with the underlying support surface. The SPF is trained in a semi-self-supervised manner and takes as an input a LiDAR pointcloud and RGB image. The network predicts a binary segmentation mask that identifies the specific points requiring adjustment, along with estimating their corresponding depth values. To train the segmentation task, 300 distinct images are manually labeled into rigid and non-rigid terrain. The depth estimation task is trained in a self-supervised manner by utilizing the future footholds of the robot to estimate the support surface based on a Gaussian process. Our method can correctly adjust the support surface prior to interacting with the terrain and is extensively tested on the quadruped robot ANYmal. We show the qualitative benefits of SPF in natural environments for elevation mapping and traversability estimation compared to using raw sensor measurements and existing smoothing methods. Quantitative analysis is performed in various natural environments, and an improvement by 48\% RMSE is achieved within a meadow terrain.
\end{abstract}

%% file: chapters/1_introduction.tex
A comprehensive understanding of the robot's surroundings is a key element in achieving autonomous robot navigation in outdoor environments.
Typically, ground robots are equipped with exteroceptive sensors such as LiDARs and depth cameras to directly capture the structure of the 3D environment. The captured geometrical information can be accumulated in a 2D occupancy map~\cite{Eitan2010Marathon, macenski2020marathon2}, 2.5D elevation map~\cite{Taka2022Cupy,Fankhauser2018TerrainMapping}, or 3D voxel map~\cite{oleynikova2017voxblox} representation.
These maps provide information about the support surface (the rigid surface that can provide support for the robot during traversing) and obstacles for downstream tasks such as motion planning~\cite{Takahiro2022locomotion, jenelten2022tamols}, traversability assessment~\cite{Frey2022Traversability, Frey2023}, and trajectory planning~\cite{oleynikova2018voxelplanning,dang2020gbplanner}.
\begin{figure}[t]
    \centering
    \includegraphics[width=\linewidth,trim=2cm 1cm 1cm 1cm,clip]{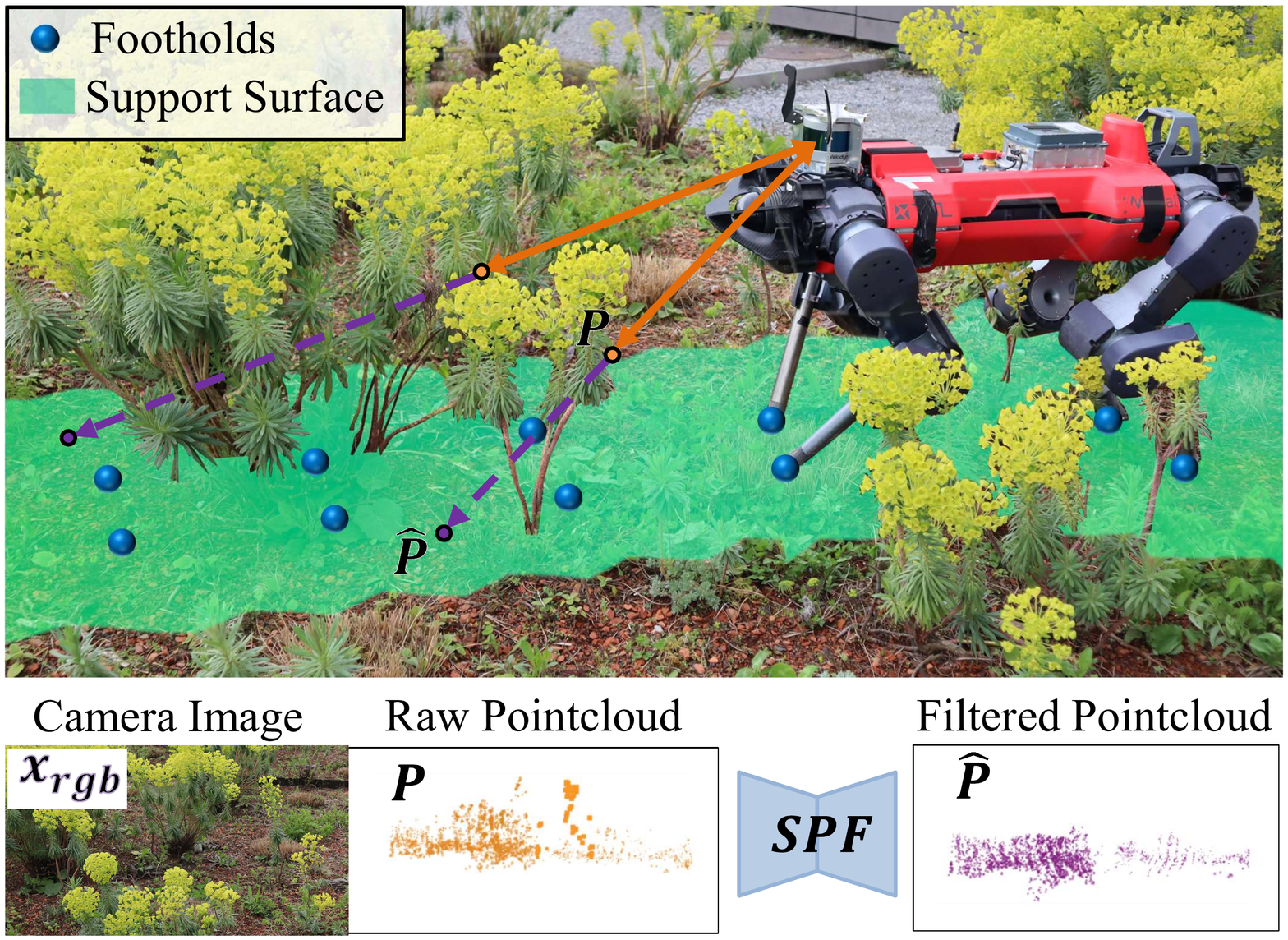}
    \caption{A systematic sketch for Semantic Pointcloud Filter (SPF). From its onboard LiDAR sensors, the raw pointcloud $P$ (orange) is generated. The raw pointcloud captures the structure of vegetation, which is undesirable for the reconstruction of the support surface. Utilizing the semantics extracted from onboard sensors, our proposed SPF filters the pointcloud by adjusting the depth of each point in the cloud. The filtered pointcloud $\hat{P}$ (magenta) is flatter and depicts the support surface.}
    \label{fig:concept}
    \vspace{-0.5cm}
\end{figure}
However, these conventional approaches are based on the assumption that the world is rigid, i.e., the robot will step on the terrain or collide with the obstacles rather than moving or deforming them. While the assumption holds in most structured environments, it is not the case for unstructured natural environments with vegetation or soft terrain. 
Most navigation planning and control pipelines often treat the penetrable vegetation as rigid, which leads to sub-optimal locomotion and navigation.
For example, a robot may try to step over vegetation or avoid non-existing obstacles without the capability to identify the support surface.
In contrast, humans easily traverse natural terrains by associating semantic information with the terrain property.
For example, before walking on grass, humans can anticipate from visual information that the high grass is penetrable and correctly adjust their gait. 

In this work, we present a novel approach to enhance the environment perception capabilities of robots by learning the support surface from RGB images and LiDAR, thereby overcoming the limitations of prior geometric methods. 
Instead of directly acquiring a fused support surface representation, such as an elevation map or voxel map, we opt to refine raw sensor data for broader downstream application adaptability.
Specifically, we propose a multi-modal model that filters pointcloud measurements by jointly performing semantic segmentation and depth estimation.
The semantic segmentation module identifies the regions where the raw pointcloud does not align with the support surface, enabling the preservation of rigid obstacles during pointcloud filtering, which is essential for navigation tasks.
Considering the inherent challenge associated with annotating depth data, the depth estimation component employs a self-supervised learning approach by leveraging supervision signals generated by foothold positions. Conversely, the semantic segmentation component is trained in a supervised manner since the task of labeling the regions within an image that require depth adjustment is comparatively less arduous.

We deploy our method on the legged robot ANYmal and validate it in a range of outdoor environments, including Grassland, Forest, and Hillside, against existing methods.
Additionally, we show the benefit of our method for multiple robotic downstream tasks, including elevation mapping and traversability estimation. Our results highlight our approach's practicality and potential applicability in real-world scenarios.

Our main contributions are the following:
\begin{enumerate}
    \item A novel Semantic Pointcloud Filter architecture that enables the filtering of raw LiDAR data to achieve alignment with the underlying support surface.
    \item A semi-self-supervised learning framework that jointly performs support surface segmentation and depth estimation. Our framework utilizes manually labeled segmentation masks and support surface depths annotated by foothold positions.
    \item Real-world experiments using the legged robot ANYmal, showcasing the benefits of the \gls{spf} for elevation mapping and traversability estimation.
\end{enumerate}

%% file: chapters/2_related_work.tex
We provide an overview of existing support surface estimation methods and the building blocks of our work, namely, depth estimation and semantic segmentation.



\subsection{Support Surface Estimation}
Support surface estimation focuses on determining stable surfaces for robot mobility. 
In the situation where e.g. sparse vegetation occludes the support surface, the vegetation can be filtered out by regarding them outliers.
This problem can be tackled by methods using Kalman Filters~\cite{Fankhauser2018TerrainMapping, thrun2002robotic, Fankhauser2014ROBOTCENTRICEM} and other heuristics~\cite{jenelten2022tamols, Kolter2009Stereovision}. 


However, these methods assume that the underlying surface is mostly flat. One can get a more realistic and robust estimation of the support surface by explicitly modeling the vegetation and the support surface. Various probabilistic models capturing both vegetation and ground characteristics~\cite{Wellington2006AGM, Homberger2019SupportSurfaceEstimation} have been proposed. They use proprioceptive and exteroceptive information to estimate support surfaces.

Rather than using heuristic modeling, Šalanský~et~al.~\cite{salansky2021KKT} introduced a learning-based method to filter a 2.5D map, while 
Wellington~et~al.~\cite{wellington2006learning} employ a learning approach to filter a voxel map.
In our research, instead of using a fused representation such as an elevation map or voxel map, we refine the raw sensor measurements directly. This allows our approach to generalize more effectively for a wider range of downstream applications.
  
\subsection{Self-supervised Depth Estimation}

Supervised depth estimation has a significant body of work~\cite{depthRes, depthDia}, but requires pixel-wise depth labeling. Self-supervised learning has been applied to monocular depth estimation, achieving better performance than supervised methods~\cite{Zhou2021SelfSupervisedMD, Johnston2020SelfSupervisedMT}. Further research incorporates semantic information~\cite{Klingner2020SelfSupervisedMD, Kumar2021SynDistNetSM}, but monocular depth estimation remains challenging due to its ill-posed nature.

Depth completion methods use LiDAR pointcloud as additional input to provide explicit geometric information~\cite{Ma2019Sparse-to-Dense, Liu2021LearningSK}. The trained networks often employ an encoder-decoder structure~\cite{Ronneberger-2015-U-Net}, which we also adapt in our work to estimate the depth of the support surface. A similar problem of filtering a point cloud in the domain of city-scale mapping is tackled by Stucker et al.~\cite{Stucker2018SUPERVISEDOD} where they leverage semantics to refine a large pointcloud scan.
\subsection{Multi-modal Semantic Segmentation}
Image-based semantic segmentation has been successfully applied to various robotic applications, including terrain classification ~\cite{valada2018deep, kuang2022semantic}.
To further improve the precision, various studies have explored the fusion of RGB and depth data for multi-modal semantic segmentation \cite{technologies10040090, HyperDense}. Typically, these methods employ convolutional neural networks (CNNs) to process input data and fuse features at different stages of the network. 
In our task, we employ semantic segmentation as a means of aiding the estimation of the depth of the support surface.

%% file: chapters/3_method.tex
The overview of our approach is given in Fig.~\ref{fig:system}.
To better distinguish the supervisions for two tasks, the self-supervised label for depth estimation is named the \gls{ssde} label, and the label obtained manually for semantic segmentation is named \gls{ssseg} label. 
The whole process comprises three stages: \gls{ssde} labels for depth estimation are generated from robot trajectories in a self-supervised manner without the need for human annotation. Subsequently, the \gls{ssseg} labels for semantic segmentation are manually labeled. Then the \gls{spf} is trained to do joint depth estimation and semantic segmentation. Finally, the pointcloud predicted by the \gls{spf} can be used for downstream applications.


\subsection{Problem Definition}
\label{ssec:problem_defn}


As shown in Fig.~\ref{fig:concept}, given a camera image $\mathbf{x}_{rgb}\in \mathbb{R}^{m\times n \times 3}$ and raw pointcloud $P_i \in \mathbb{R}^3, i\in \{1\dots N\}$, our objective is to derive the filtered pointcloud, $\hat{P}$, using our SPF, $\hat{P} = f_\mathrm{SPF}(P, \mathbf{x}_{rgb})$. 
Each point in the filtered pointcloud, $\hat{P}_i\in \mathbb{R}^3, i\in \{1\dots N\}$, aligns with the support surfaces or the impenetrable obstacles.


\subsection{\gls{ssde} Label Generation}
\label{ssec:label-gen}
We extract the foothold positions of the robot and approximate the support surface using a \gls{GP} similar to~\cite{Homberger2019SupportSurfaceEstimation}. 
Then we generate pixel-wise depth labels by projecting the reconstructed support surface into the image.

\subsubsection{Foothold extraction}
Foothold positions serve as ground truth samples for the support surface. From the dataset, foot position trajectories are obtained using forward kinematics, comprising discrete height points over time. Footholds are extracted, assuming foot-ground contact occurs at local height minima in the \textit{world frame}. 

\begin{figure}[!htp]
    \centering
    \includegraphics[width=\linewidth,trim=3.5cm 2.8cm 6cm 0.5cm,clip]{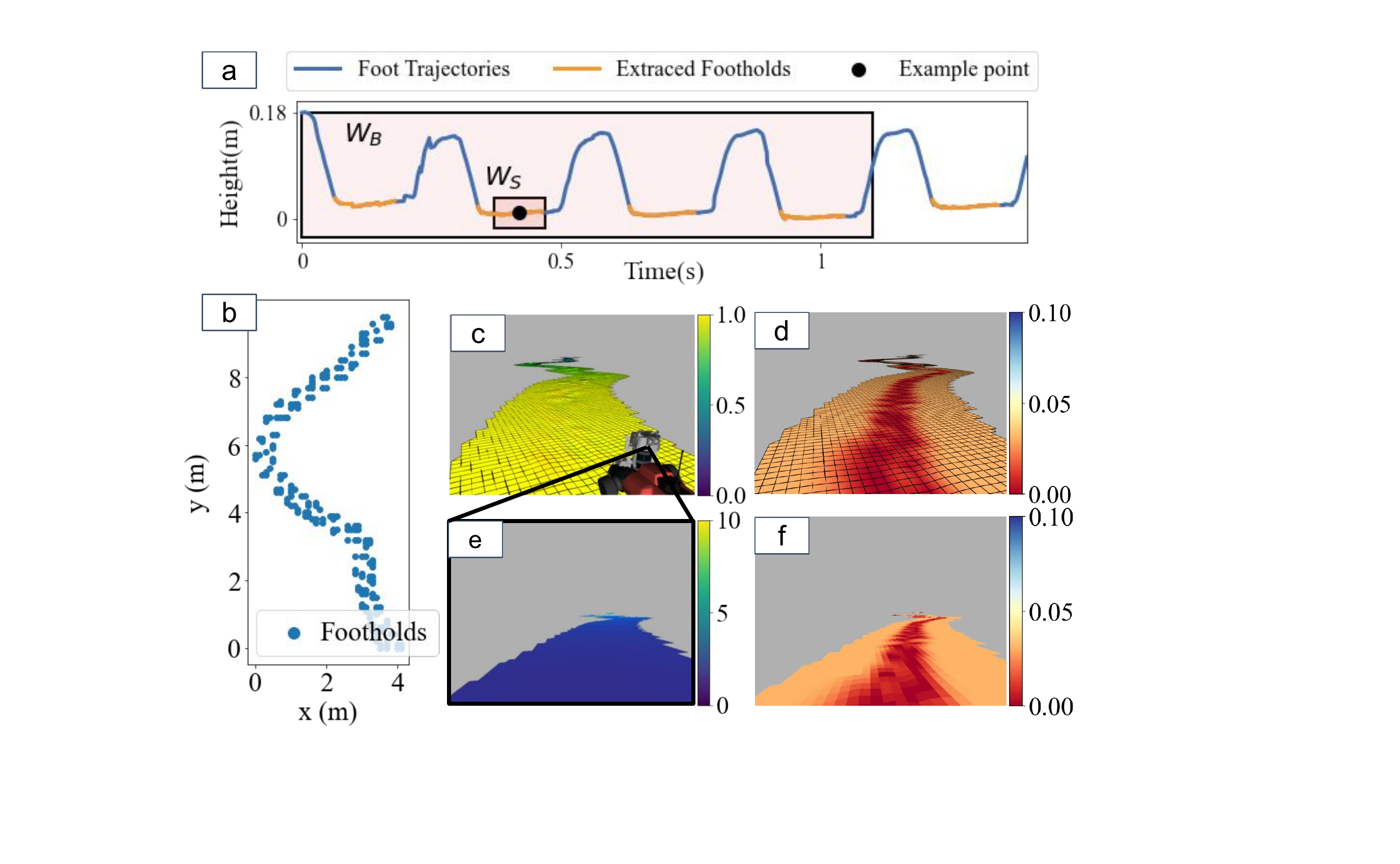}
    \caption{The reconstruction of the support surface. (a) The method to extract footholds. (b) Top view of the sparse grid map with the extracted foothold positions.  (c) Reconstructed support surface using Gaussian Processes (GP) with color-coded elevation. (d) The corresponding variance of the reconstruction. The red area is with lower variance than the orange area. (e) The sparse depth image of the reconstructed support surface with the distance from the camera focal point being color-coded. (f) The corresponding variance of e in the image plane.}
    \label{fig:extract}
    \vspace{-0.3cm}
\end{figure}

Specifically, to extract footholds, a large temporal window $W_B$ with duration $t_B$ is constructed, and a short temporal window $W_S$ with duration $t_S$ is built around a point within $W_B$ (Fig.~\ref{fig:extract}a). Mean foot heights within $W_S$ and $W_B$ are calculated, yielding $mean_S$ and $mean_B$, and the maximum height change within $W_S$ is denoted as $r_S$. A point is identified as a foothold if $mean_S \leq mean_B$ and $r_S \leq r_{t}$, where $r_t$ is a threshold. This is applied iteratively for each point in $W_B$. After evaluating all points, $W_B$ is moved forward, and the process is repeated. Footholds are denoted as $P_{i} \in \mathbb{R}^3, i\in {1\dots M}$, with $M$ as the number of extracted footholds.

\subsubsection{Support surface reconstruction}
\label{sssec:GP}


We extrapolate the positions of the extracted footholds in \textit{world frame} to generate the support surface as done by Homberger et al.~\cite{Homberger2019SupportSurfaceEstimation}. The foothold extrapolation method is valid under the assumption that the support surface is continuous and smooth.

To achieve this, a grid map is employed, and the sparsely distributed footholds are used to determine the height of the corresponding cells in the grid map (Fig.~\ref{fig:extract}b). This sparse grid map is then utilized to extrapolate the height values to adjacent grid cells via the implementation of a \gls{GP}.

Given a group of footholds $P_{i} = (\mathbf{x}_i, z_i)$, we stack their x-y positions into $\mathbf{X}= (\textbf{x}_1, ... , \textbf{x}_M)$ and heights  $\mathbf{Z}= (z_1, ... , z_M)$. We model them as a \gls{GP} with a data distribution of 
\begin{equation}
P\left(\mathbf{Z} \mid \mathbf{X} \right) \sim \mathcal{N}(\mathbf{\mu}, \mathbf{K} + \sigma_n^2 * \mathbf{I})
\end{equation}
with $\mu \in \mathbb{R}^n$ the mean of $\mathbf{Z}$, $\sigma _n$ the noise and K the covariance matrix, $K_{i, j} = k(\mathbf{x}_i, \mathbf{x}_j)$, where k is the kernel function~\cite{williams2006gaussian}. Then the distribution of the prediction at position $\mathbf{x}^*$ can be written as $z ^* \sim \mathcal{N}(\mu ^*, v^*)$, where

\begin{equation}
\begin{gathered}
\mu^*=\mathbf{k}^T\left(\mathbf{K}+\sigma_n^2 \mathbf{I}\right)^{-1} \mathbf{Z} \\
v^*=k^*+\sigma_n^2-\mathbf{k}^T\left(\mathbf{K}+\sigma_n^2 \mathbf{I}\right)^{-1} \mathbf{k}
\end{gathered}
\end{equation}
with $\mathbf{k} \in \mathbb{R}^n$, $k_j = k(\mathbf{x}^*, \mathbf{x}_j)$. 
Our experimental results indicate that employing solely the \gls{RBF} kernel is sufficient to achieve a high level of accuracy and smoothness in the reconstructed support surface:
\begin{equation}
\label{eq:GP_kernel}
    k\left(\mathbf{x}_i, \mathbf{x}_j\right)= \exp \left(-\frac{\left\|\mathbf{x}_i-\mathbf{x}_j\right\|^2}{2 l^2}\right)
\end{equation}

To reduce the computational complexity, we fit multiple \gls{GP}s to small overlapping tiles for each local grid map. Subsequently, the individual \gls{GP}s are integrated into a composite grid map that provides a variance and height estimate to represent the support surface. The height value is determined by calculating the mean \gls{GP} prediction for the overlapping regions, while the variance is set to the maximum variance.

\subsubsection{Support Surface Projection}
\label{sssec: Labels generation}
To generate training labels for \gls{spf} depth estimation, we reproject the generated support surface (Sec.~\ref{sssec:GP}) into all captured camera images along the trajectory. 
By raycasting the generated height grid map (Fig.~\ref{fig:extract}c), we obtain correspondences between the pixels on the image plane and grid map cells. 
Given the correspondences, we can generate a support surface depth image capturing the distance from the camera center to the support surface (Fig.~\ref{fig:extract}e), as well as project the variance of the Gaussian process $\sigma$ into the image plane (Fig.~\ref{fig:extract}f). The variance of the Gaussian Process does not capture the uncertainty with respect to the distance from which the support surface is observed. We found experimentally that using a more sophisticated formulation incorporating the observation distance during training did not increase the performance.
The variance is used to mask the region of valid depth information using a threshold parameter $\tau_{max}$, resulting in the support surface label $\mathbf{y}_{{de}}$ used as a target during training.

\subsection{Semantic Pointcloud Filter}
\label{ssec:method-spf}
To leverage image-based computer vision techniques, $P$ is projected to the camera image plane and obtain a sparse depth image $\mathbf{x}_{d}\in \mathbb{R}^{m\times n}$. 
We use a image processing neural network to predict the filtered depth, $\hat{\mathbf{y}}=h(\mathbf{x}_{rgb}, \mathbf{x}_{d})$. 
The filtered depth image is re-projected to the space as the filtered pointcloud $\hat{P}$.
\subsubsection{Network Structure}
The neural network model follows an encoder-decoder structure with skip connections, as proposed in~\cite{Ronneberger-2015-U-Net}. 
The encoder employs a pretrained EfficientNet-B5~\cite{tan2019efficientnet} as the backbone, where the first convolution layer is adapted to accommodate 4 input channels. 
The input to the network is a stacked tensor of the sparse depth image $\mathbf{x}_{d}$ and the camera image $\mathbf{x}_{rgb}$. 

The decoder concurrently generates the support surface depth $\hat{\mathbf{y}}_{de}$, and binary semantic segmentation mask $\hat{\mathbf{y}}_{s}$ with two classes: \textit{rigid obstacles} and \textit{support surfce}. 
The final output $\hat{\mathbf{y}}$ combines  the support surface depth estimation $\hat{\mathbf{y}}_{de}$ with the raw pointcloud $\mathbf{x}_{d}$
\begin{equation}
    \hat{\mathbf{y}} = \mathbf{x}_{d}[ \hat{\mathbf{y}}_{s} == rigid]  + \hat{\mathbf{y}}_{de}[\hat{\mathbf{y}}_{s} ==support ]
\end{equation}

\subsubsection{Training}
\label{subsub:train}
The network is trained to minimize the total training loss $\mathcal{L}_{\mathrm{total}}$, given by the weighted sum of the support surface depth estimation loss $\mathcal{L}_{\mathrm{de}}$ and semantic segmentation loss $\mathcal{L}_{\mathrm{seg}}$:


\begin{equation}
\label{eq:loss}
\mathcal{L_{\mathrm{total}}}=  \mathcal{L}_{\mathrm{seg}} + w\mathcal{L}_{\mathrm{de}}
\end{equation}
where $w$ is used to balance the loss contributed by depth estimation and semantic segmentation. $\mathcal{L}_{\mathrm{de}}$ of prediction $\hat{\mathbf{y}}_{\mathrm{de}}$ is computed as the mean squared error (MSE) with respect to \gls{ssde} labels, and is weighted by the pixel-wise variance $\sigma_i$ of the support surface. 
\begin{equation}. 
\mathcal{L}_{\mathrm{ss}} =  \sum_{i\in C_{\mathrm{ss}}} \sigma_{i}^{-2}\left(\hat{\mathbf{y}}_{de,i}-\mathbf{y}_{\mathrm{de},i}\right)^{2}
\end{equation}
 $C_{\mathrm{ss}}$ is set of pixels where $\mathbf{y}_{de}$ has valid values. 
 While $\mathcal{L}_{\mathrm{seg}}$ is computed by employing the cross-entropy loss with respect to \gls{ssseg} labels, which is commonly used for semantic segmentation tasks. 
 

%% file: chapters/4__Implementation_Details.tex
\subsection{\gls{ssde} Generation}
When generating the \gls{ssde} labels, we set the length of $W_B$ to \SI{1.5}{s}, length of $S_B$ to \SI{0.07}{s} and $r_t$ to \SI{0.015}{m}. 
In RBF kernel, the $l$ is set to $1e^{-5}$.
For the depth image of the reconstructed terrain, we only use the region with a variance lower than $\tau_{max}=0.03$ as our \gls{ssde} labels. 

\subsection{Network Training}
The neural network is trained for 9000 steps ($\sim$\SI{1}{hour}) using a batch size of 6 on a single NVIDIA GeForce RTX3090, with an EfficientNet-B5~\cite{tan2019efficientnet} encoder pre-trained on ImageNet~\cite{5206848}. 
The resulting weights of the additional depth input channel and the decoder are randomly initialized following~\cite{glorot2010understanding}. To train the network, we use the AdamW optimizer~\cite{loshchilov2017decoupled} and schedule the learning rate following~\cite{Farooq2021AdaBins}. In the loss function, the $w$ is set to 0.02.
During training for data augmentations, we first flip $\mathbf{x}_{rgb}$ and $\mathbf{x}_{d}$ horizontally with a probability of 0.5. Secondly, the input images are randomly cropped from a shape of 540 $\times$ 720 to 352 $\times$ 704.

%% file: chapters/4_experiments.tex
We first describe the experiment dataset collection and training data generation (Sec.~\ref{ssec:dataset-overview}). 
Then we test the trained \gls{spf} depth prediction performance and perform an ablation study (Sec.~\ref{ssec:ablation-study}). 
Finally, we showcase that the \gls{spf} filtered pointcloud can generate more accurate results in two downstream applications, namely elevation mapping~(Sec.~\ref{ssec:res_elevation_map}) and traversability estimation (Sec.~\ref{ssec:res-traversability-estima}).

\subsection{Dataset Overview}
\label{ssec:dataset-overview}
\begin{figure}[htp]
    \centering
    \includegraphics[width=\linewidth]{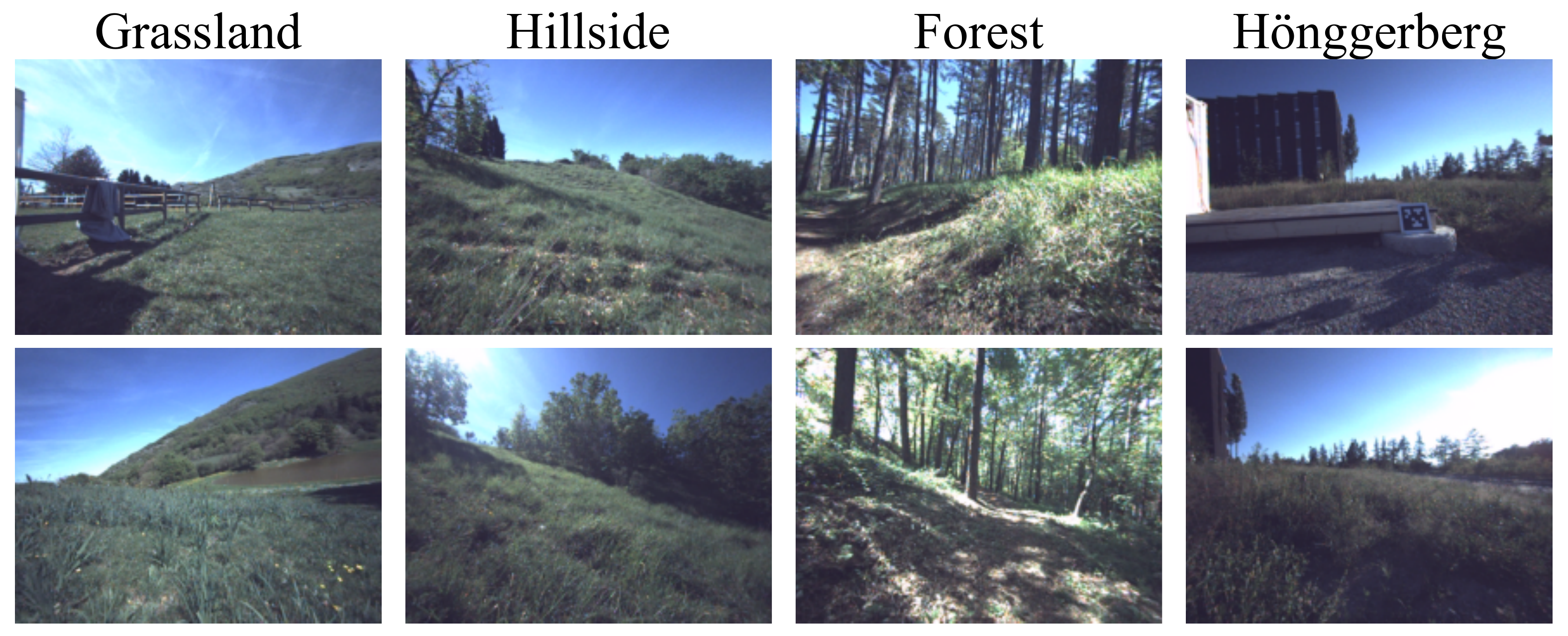}
    \caption{Example images from different environments.}
    \label{fig:Image_examples}
    \vspace{-0.3cm}
\end{figure}
The data is collected using the legged robot ANYmal, controlled by a human operator, in various outdoor environments in Perugia, Italy and Hönggerberg, Switzerland. 
ANYmal is equipped with a Robosense RS-Bpearl 32 beam Dome-LiDAR and an Alphasense Core \SI{0.4}{MPix} RGB camera unit. The camera images and LiDAR data are collected at a rate of \SI{10}{Hz}, and downsampled during post-processing to \SI{2}{Hz}, with image resolution at 540 $\times$ 720. 
We recorded six 10-20 minutes trajectories from Perugia, classified into three environments: Grassland, Hillside, and Forest. Each environment contains two trajectories. The training set and testing set each contain three trajectories, one from each of these three environments.
For each trajectory, 100 images were sampled at a 3.5-second interval to avoid repetition, and each image was manually annotated into two classes: \textit{support surface} and \textit{rigid obstacles} for semantic segmentation.
Additionally, a trajectory was collected from Hönggerberg, Switzerland, only for evaluating downstream applications.

Examples of images from the onboard camera are shown for the different environments in Fig.~\ref{fig:Image_examples}. 

\subsection{Support Surface Depth Estimation}
\begin{figure}[htp]
    \centering
    \includegraphics[width=\linewidth]{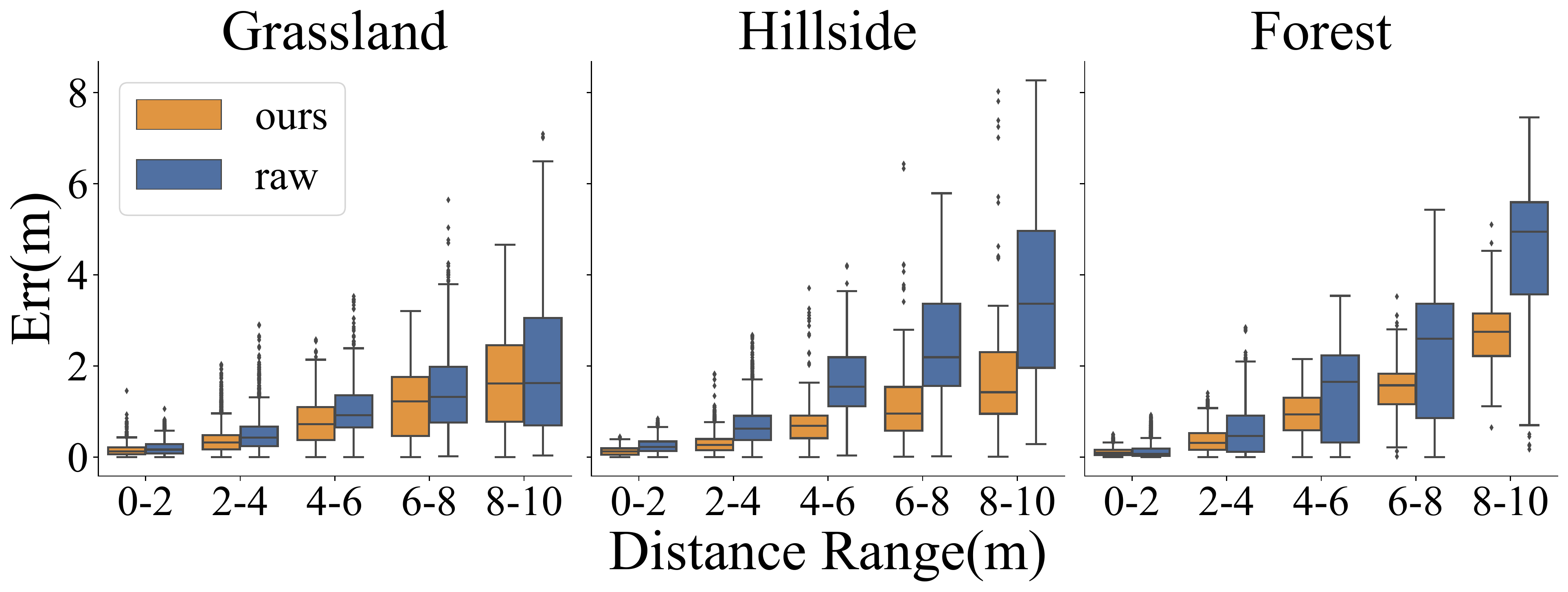}
    \caption{Comparision of the absolute error of the depth estimation with respect to the distance in the different environments in the testing set.}
    \label{fig:evaluation_depth}
\end{figure}

The result presented in Fig.~\ref{fig:evaluation_depth} shows the distribution of the absolute error of support surface depth estimation in relation to distance compared to the raw pointcloud. 
The absolute error is computed with respect to the \gls{ssde} labels from the camera and is binned by the distance. 
The result demonstrates that the performance of the raw pointcloud worsens in Grassland, Hillside, and Forest environments. The filtered pointcloud also exhibits this trend, indicating that the quality of the raw pointcloud directly affects the performance of our filtered pointcloud. Notably, in all environments, within any distance bins, the filtered pointcloud outperforms the raw pointcloud.
The error increases with respect to the distance, which is common in monocular depth estimation tasks. Additionally, the reconstructed support surface, while being locally consistent in proximity to the robot, becomes less accurate at longer ranges due to localization drift.

\subsection{Ablation Study}
\label{ssec:ablation-study}

\begin{table*}[htp]
	\centering
	\ra{1.2}
	\footnotesize
	\setlength{\tabcolsep}{3pt}
 	\caption{Evaluation of Support Surface Depth Estimation.}
    \begin{tabular}{lcclcclcclcclcc} \toprule
        & \multicolumn{2}{c}{Grassland } && \multicolumn{2}{c}{Hillside}  && \multicolumn{2}{c}{Forest} && \multicolumn{2}{c}{All}\\
        \cmidrule{2-3} \cmidrule{5-6} \cmidrule{8-9} \cmidrule{11-12} 
       Method & RMSE$\downarrow$ & REL$\downarrow$  && RMSE$\downarrow$ & REL$\downarrow$ && RMSE$\downarrow$ & REL$\downarrow$ && RMSE$\downarrow$ & REL$\downarrow$ \\
       
        \midrule 
        Raw Pointcloud & 0.503 & 0.15 &&  0.710 & 0.20  && 0.656 & 0.15 &&0.63& 0.17\\ 
        Only RGB & 0.46 $\pm$ 0.03 & 0.15 $\pm$ 0.06 && 0.37 $\pm$ 0.03 & 0.11 $\pm$ 0.04 && \textbf{0.35 $\pm$ 0.03} & \textbf{0.09 $\pm$ 0.02} && 0.40 $\pm$ 0.03 & 0.12 $\pm$ 0.03\\
        Only LiDAR & 0.45 $\pm$ 0.03 & 0.11 $\pm$ 0.01  && \textbf{0.29 $\pm$ 0.02} & \textbf{0.08 $\pm$ 0.01} && 0.46 $\pm$ 0.04 & 0.10 $\pm$ 0.01 && 0.41 $\pm$ 0.01 & 0.10 $\pm$ 0.03\\

    SPF & \textbf{0.37 $\pm$ 0.02} & \textbf{0.11 $\pm$ 0.01} && 0.31 $\pm$ 0.04 & 0.10 $\pm$ 0.01 && 0.37 $\pm$ 0.02 & 0.10 $\pm$ 0.01 && \textbf{0.35 $\pm$ 0.01} & \textbf{0.10 $\pm$ 0.01}\\
    \bottomrule
    \end{tabular}
\label{tab: evaluation_de}
\end{table*}

\begin{table}[htp]
	\centering
	\ra{1.2}
	\footnotesize
	\setlength{\tabcolsep}{2.4pt}
 	\caption{Support Surface Semantic Segmentation in mIoU.}
    \begin{tabular}{lcclclclclc} \toprule
        & \multicolumn{1}{c}{Grassland} && \multicolumn{1}{c}{Hillside}  && \multicolumn{1}{c}{Forest} && \multicolumn{1}{c}{All}\\
       
        \midrule 
        No hand-label & 43.93$\pm$0.93&&48.61$\pm$0.96&&50.26$\pm$1.19&&45.99$\pm$0.68\\
        Only RGB & 89.12$\pm$1.33&&\textbf{83.71$\pm$2.67}&&96.90$\pm$0.85&&89.02$\pm$0.72\\
        Only LiDAR & \textbf{90.77$\pm$0.21}&&83.28$\pm$0.89&&96.50$\pm$0.07&&89.14$\pm$0.33\\
    SPF & 90.46$\pm$0.35&&82.83$\pm$0.92&&\textbf{97.29$\pm$0.17}&&\textbf{89.42$\pm$0.38}\\

    \bottomrule
    \end{tabular}
\label{tab: evaluation_ss}
\vspace{-0.3cm}
\end{table}

The first ablation assesses the impact of providing varying input modalities followed by investigating the use of hand-labeled \gls{ssseg} labels in the training of the semantic segmentation task. 
For rigors analysis, we train all models 10 times with varying random seeds and report the standard deviation.

We evaluate depth estimation and semantic segmentation separately for each model. To measure the performance of depth estimation, we use the following metrics:
\begin{enumerate}
    \item \hspace*{-0.2cm} Average Relative Error (REL): $\frac{1}{n} \sum_{p}^{n} \frac{\left|y_{p}-\hat{y}_{p}\right|}{y}$
    \item \hspace*{-0.2cm} Root Mean Squared Error (RMSE): $\sqrt{\left.\frac{1}{n} \sum_{p}^{n}\left(y_{p}-\hat{y}_{p}\right)^{2}\right)}$
\end{enumerate}
where all metrics are commonly used to evaluate monocular depth estimation performance~\cite{Farooq2021AdaBins}. The metrics are evaluated with respect to all the validation pixels of the SSDE label within 5m. The result is shown in Table~\ref{tab: evaluation_de}. It is important to note that the error in depth estimation is assessed within the image space, and therefore, does not directly correspond to the error in the estimation of the height of the support surface.

\begin{figure}[htp]
    \centering
    \includegraphics[width=\linewidth]{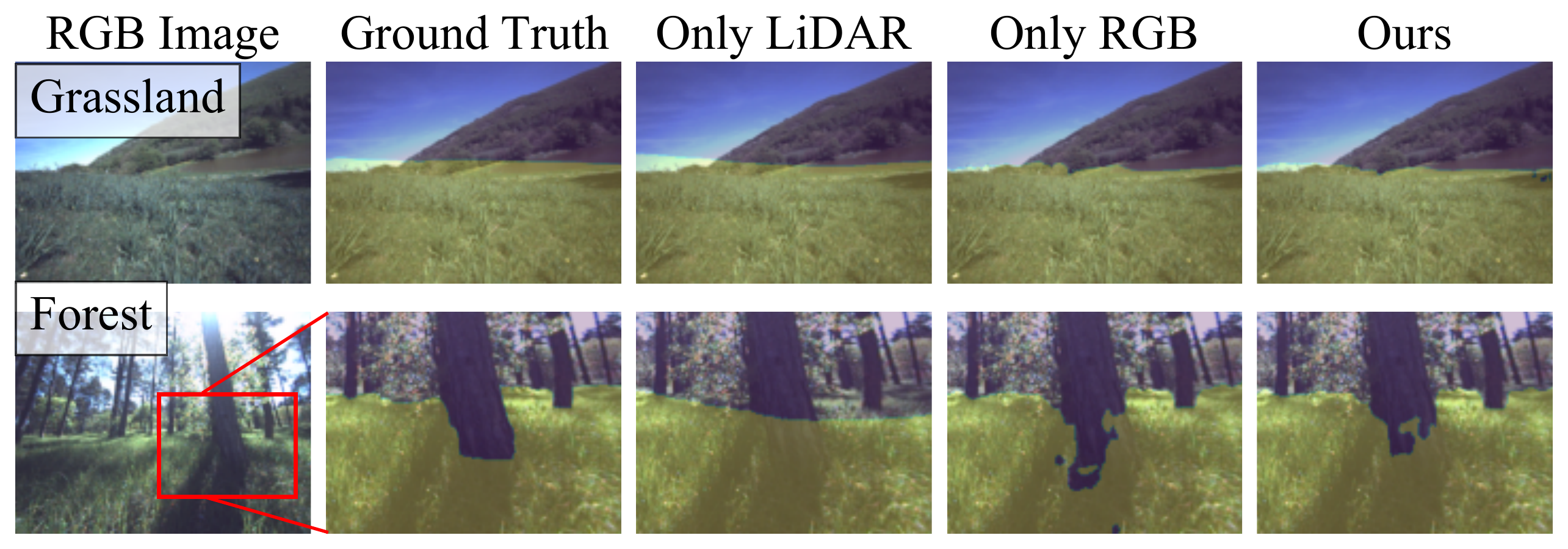}
    \caption{The impact of varying modality input on semantic segmentation. 
    In a relatively simple environment with no rigid obstacles, the segmentation performance of all examined models appears to be comparable (1st row). 
    However, in more complex environments (2nd row), the model relying solely on pointcloud input struggles to accurately segment the \textit{rigid obstacles} near the ground (the trunk), while the model utilizing only RGB images is adversely impacted by the tree's shadow. 
    }
    \label{fig:evaluation_ss}
    \vspace{-0.3cm}
\end{figure}


We present an ablation comparison for support surface depth estimation in Table~\ref{tab: evaluation_de}. All our models, as well as the \textit{Only RGB} and \textit{Only LiDAR} input models, outperform the raw pointcloud. 
In average, our model outperforms single-modality models (\textit{Only RGB}, \textit{Only LiDAR}) by 14\% in terms of RMSE. 
While single-modal models may yield better results in specific environments, our model remains competitive and close (less than \SI{0.02}{m}) to the best ablations in all environments. 
Conversely, ablation models can exhibit an RMSE increase of more than 20\% in comparison to our model in less favorable environments. 
These findings indicate that the multi-modal design significantly improves the model's robustness in the face of varying environmental conditions.


We evaluate the semantic segmentation of ours and the ablated models with the \gls{miou} metric, which is a widely adopted measurement in the field of semantic segmentation tasks~\cite{ConvolutionForSegmentation}. 
We show the results for two exemplary inputs in Fig.~\ref{fig:evaluation_ss}, and provide numerical \gls{miou} values in Table~\ref{tab: evaluation_ss}.
Surprisingly, our results in Table.~\ref{tab: evaluation_ss} indicate that our model, as well as two ablations, demonstrate comparable performance in \gls{miou}. 
This can be attributed to two factors.
First, as illustrated in Fig.~\ref{fig:evaluation_ss}, the majority of the image is relatively straightforward to segment. The key challenge lies in accurately determining the boundaries of the support surface, which represents only a minor portion in comparison to the entire label. 
This causes the \gls{miou} to be less sensitive to the segmentation of the area of interest.
Secondly, while RGB images provide more explicit semantic information than pointcloud, they can suffer from poor lighting conditions (as demonstrated in Fig.~\ref{fig:evaluation_ss}), making it difficult to generalize on our small dataset.
Consequently, incorporating geometric information into our model can potentially enhance its robustness.
Considering all three environments, our ablation shows that multi-modal and single-modal perform on par for the task of semantic segmentation.

We also investigate the necessity of manually labeled segmentation labels by comparing it to solely employing labels generated based on the valid regions of the \gls{ssde} label $C_{\mathrm{ss}}$ and considering other regions as rigid obstacles. This label does not reflect the scene well given that all untraversed regions are regarded as obstacles/ rigid. 
As shown in Table~\ref{tab: evaluation_ss}, the models with hand-labeled data exhibit significant improvement to the model trained without hand-labels.

\begin{figure*}[htp]
    \centering
    \includegraphics[width=\linewidth]{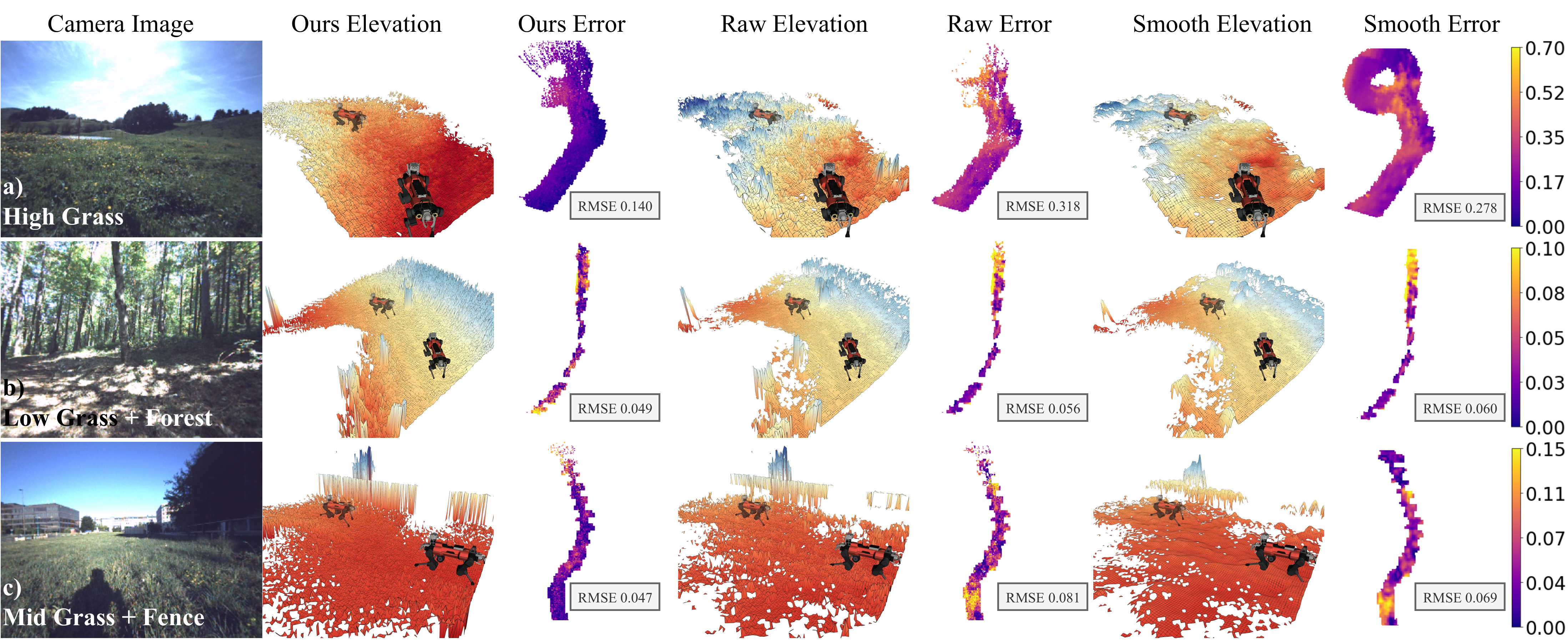}
    \caption{Comparison of elevation mapping generated from different methods in different environments. On the leftmost side are the onboard camera images. For each method, we provide the generated elevation map and the error map (RMSE) from a top-down perspective with respect to the support surface for this region of the trajectory. We compare our method (\textit{Ours})  to the raw pointcloud (\textit{Raw}) and the raw pointcloud with additional smoothing (\textit{Smooth}). The colorbars for error maps are placed on the right, with brighter colors indicating higher error. Our method outperforms the baselines in various environments, such as high grass hills (a), forests (b), and urban grasslands with impenetrable fences (c).}
    \label{fig:ElevationCompare}
    \vspace{-0.3cm}
\end{figure*}
\subsection{Elevation Mapping}
\label{ssec:res_elevation_map}

\begin{figure}[htp]
    \centering
    \includegraphics[width=\linewidth]{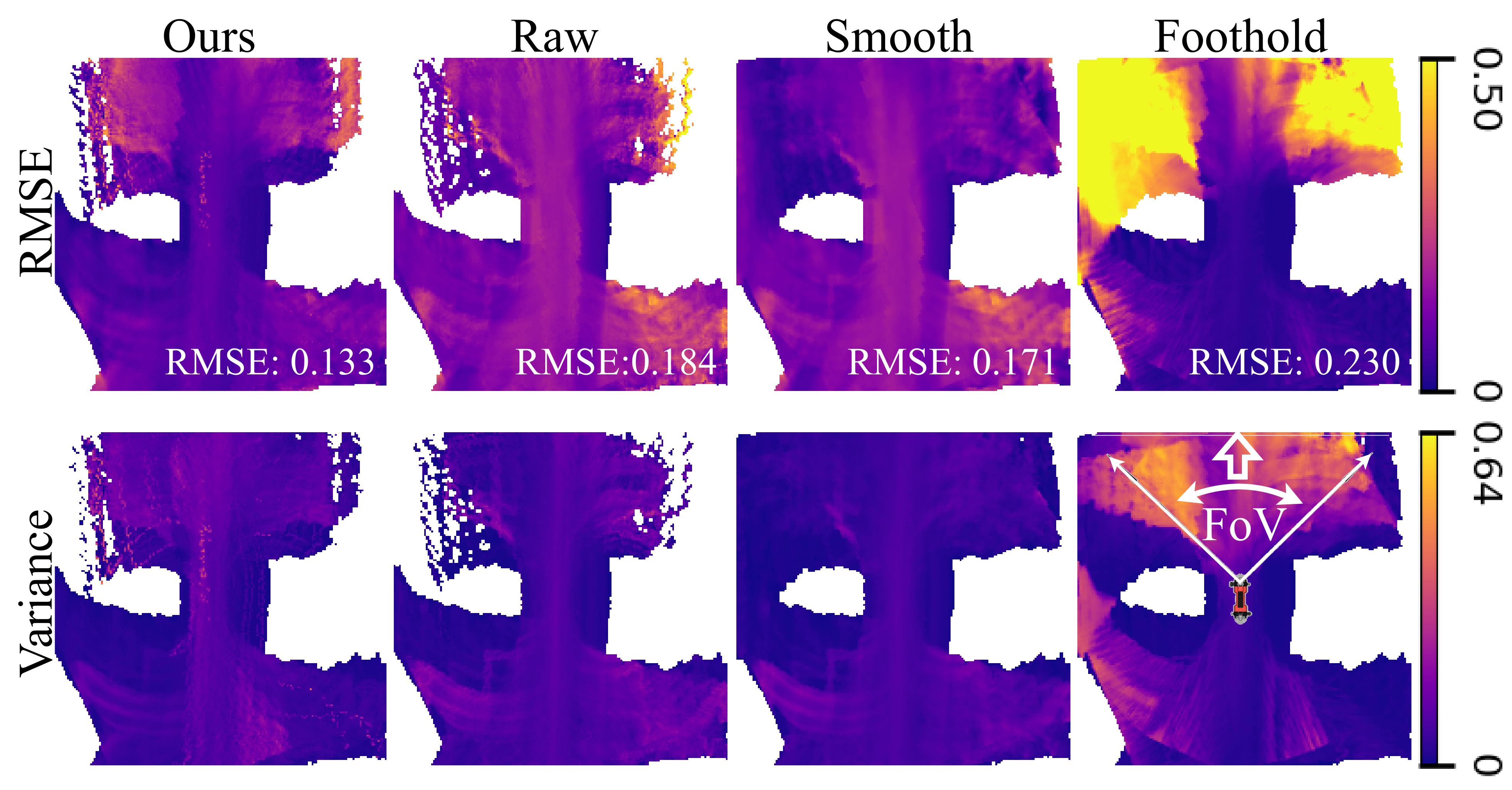}
    \caption{The errors between elevation maps generated using our method and baselines. In the first row, the value of each pixel is the \gls{rmse} of that pixel over the testing trajectory, expressed in meters. The variances (second row) of the errors are also provided. These figures are accumulated in the robot frame. The forward direction and the field of view (FoV) of the robot are shown in the bottom right figure. White areas indicated that no errors are collected in those cells.}
    \label{fig:ElevationErrorMap}
    \testvspace
\end{figure}

To validate \gls{spf}'s practical benefits, we input the filtered pointcloud into an elevation mapping algorithm~\cite{Taka2022Cupy}, which fuses the pointcloud with the robot pose to generate an \SI{8}{m}$\times$\SI{8}{m} grid map at \SI{0.04}{m} resolution.
We compare our method with three baselines:
\textit{Raw} generates maps using the same elevation mapping algorithm but inputs the raw pointcloud. \textit{Smooth} involves applying a smooth operation~\cite{Taka2022Cupy} on the elevation map generated from \textit{Raw}. 
For these two baselines, we only feed the points within the camera's \gls{fov} into the algorithm. An illustration of FoV is presented in Fig.~\ref{fig:ElevationErrorMap}.
Moreover, \textit{Foothold} refers to the nearest-neighbor completion of the footholds, where the robot traversed. This baseline exemplifies methods relying solely on proprioceptive observation and a flat-ground assumption.

The result in Fig.~\ref{fig:ElevationCompare} shows three elevation mapping samples together with two baseline methods. 
In each image, we illustrate the robot as non-transparent (past) and transparent (future). These robots allow us to validate the elevation mapping accuracy by examining the foothold alignment with the estimated terrain surface.
Sample (a) features high grass. The elevation map constructed with our SPF matches the robot's foot height, while the maps from the two baselines are higher than the robot's leg. 
Smoothing the raw elevation map reduces its \gls{rmse} by 12.6\% while our method reduces it by 56.0\%. 
Sample (b), set in a forest with minimal grass, also demonstrates our method's superior performance in representing support surfaces and recognizing obstacles. 
Sample (c), taken from Hönggerberg Zurich, highlights a domain shift from nature to an urban setting. Similar to (a), our method reduces the RMSE by 42.0\% compared to the raw elevation map and accurately captures the fence's structure.
These comparisons demonstrate that our SPF effectively and robustly distinguishes vegetation from rigid obstacles, correctly lowering the perceived vegetation height while preserving the structure of rigid obstacles.

For quantitative comparison over the testing trajectory in Perugia Grassland, we compare the constructed elevation map with the \gls{ssde} labels described in Sec.~\ref{ssec:label-gen}. 
Specifically, elevation maps are generated at a rate of \SI{10}{Hz} as the trajectory is replayed. On each update of the maps, an error is computed between the new map and the \gls{ssde} labels. Then we transform these error maps into the robot's frame, accumulate them, and compute the \gls{rmse}, and error variance, as depicted in Fig.~\ref{fig:ElevationErrorMap}.
The trajectory is from the test set featuring a hillside. 
In the trajectory, the robot was mostly walking forward with some turning motions (The forward direction is shown in the bottom right of Fig.~\ref{fig:ElevationErrorMap}). Moreover, the elevation perceived in the front goes towards the back of the robot's frame as the robot moves forward. Therefore, the upper parts (the part in FoV) of each subfigure correspond to the "prediction" of elevation height, while the lower parts correspond to the mapping result fused over several frames. 

The elevation map constructed with our filtered pointcloud has an \gls{rmse} of \SI{0.133}{m},
which is about 48\% less than the \SI{0.183}{m} of the raw baseline. 
Smoothing results in an \gls{rmse} of \SI{0.170}{m}, which is only a 7\% improvement. 
Our method has a lower error at both the front and back of the robot. The high error value on the top left is exceptional and is caused by overexposure when the robot leaves tree shades.
As the footholds are exact samples of the support surface, foothold baseline achieves almost zero \gls{rmse} in the areas under the robot and behind the robot. However, it predicts the height of the untraversed area in the front much worse than the methods with exteroceptive observations.
The standard deviations of the errors are shown in the second row. The error from the smooth baseline is more stable, resulting in a lower standard deviation. Both our method and the raw pointcloud have comparable standard deviations for their errors.
Please note that thanks to the modularity design of our proposed SPF. It's possible to leverage the benefits of both pointcloud filtering and smoothing and obtain prediction that is more accurate and stable.

\subsection{Traversability Estimation}
\label{ssec:res-traversability-estima}
\begin{figure}[htp]
    \centering
    \includegraphics[width=\linewidth]{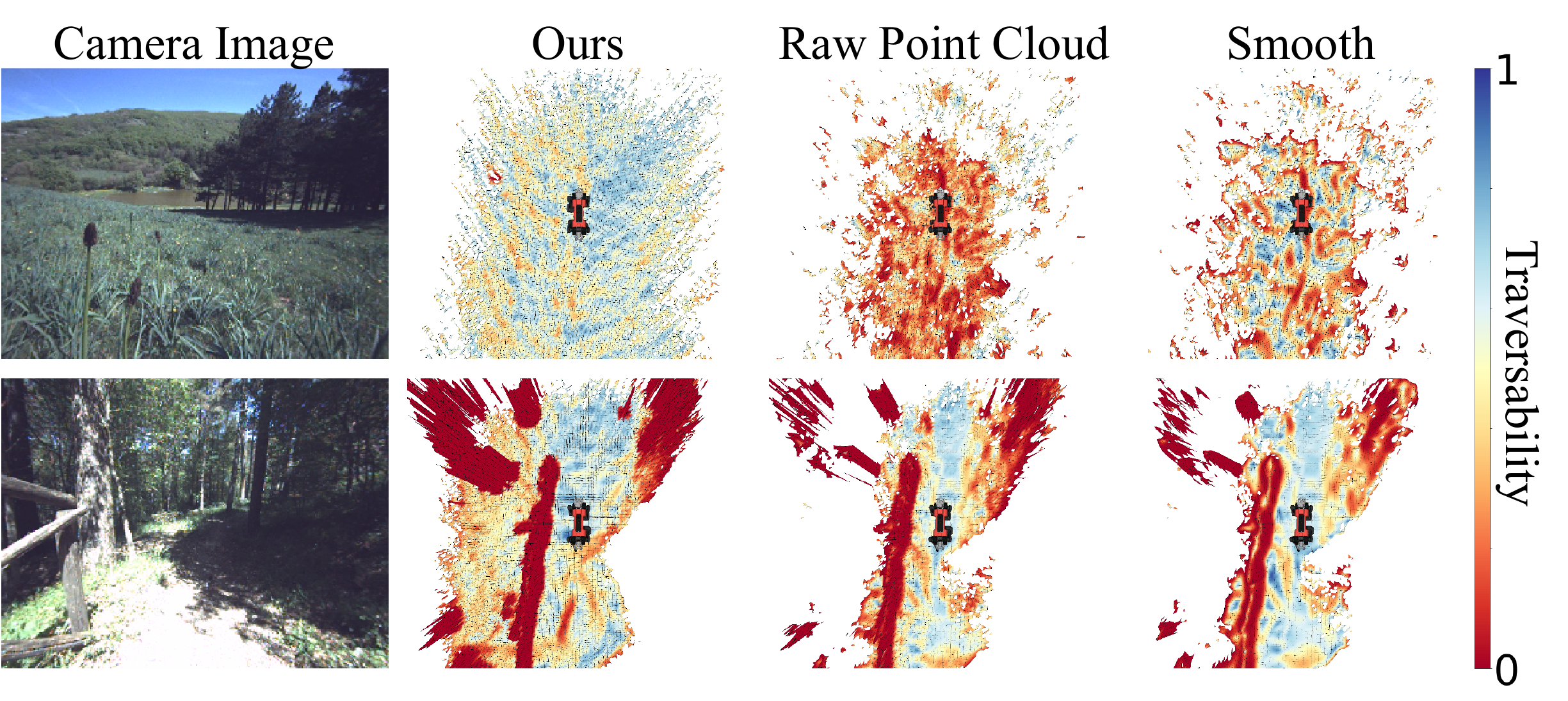}
    \caption{Comparison of traversability estimation using the raw, smoothed, and our filtered pointcloud. The traversability is color-coded, where blue indicates traversable and red untraversable. Our method correctly predicts the traversability in the meadow and forest environment. Using the raw or smoothed pointcloud prohibits motion planning in the meadow.}
    \label{fig:traversmap}
    \vspace{-0.3cm}
\end{figure}

Filtering out noisy high grass and revealing true support terrain is beneficial for an accurate traversability estimation. 
We estimate the traversability using the method described in~\cite{yang2021motioncost}. Elevation maps constructed in Sec.~\ref{ssec:res_elevation_map} are used as the input to this algorithm.
Two samples of traversability estimation in the natural environment are shown in Fig.~\ref{fig:traversmap}.
The grid maps on the right rows are colored with traversability, where a high traversability corresponds to blue, while low to red.
When the robot is in the vegetation, our method recognizes vegetation and estimates the underlying support surface leading to the correct traversability estimate.
On the other hand, directly using the pointcloud or the smoothed version  lead to incorrect traversability rendering motion planning and navigation impossible.
When the robot is in the forest, our method and baselines all yield reasonable traversability maps.

%% file: chapters/5_conclusion.tex
In this work, we present our semantic pointcloud filter (SPF) trained in a semi-self-supervised manner, which can accurately adjust the pointcloud to the support surface on a vegetation-occluded terrain, leveraging semantics from the camera images and pointcloud.
Using pointcloud filtered by SPF, we improved elevation mapping and traversability estimation performance compared to existing baseline methods, potentially increasing autonomy for a variety of robotic systems within natural environments. 

While multiple real-world experiments demonstrated the efficacy of the proposed method, limitations exist, such as errors in filtered depth estimation under adverse lighting conditions and the possibility of inaccuracy due to domain shifts in vegetation species and seasons.
Future research will explore the incorporation of anomaly detection into label generation for semantic segmentation, eliminating the need for manual labeling and scaling up data collection to enhance robustness further. Moreover, integrating the pointcloud filter with downstream tasks, including locomotion, path planning, and exploration, holds significant promise.